\documentclass[preprint,12pt]{elsarticle}

\usepackage{amssymb}
\usepackage{amsmath}

\journal{Nuclear Physics B}
\usepackage{booktabs}
\usepackage{microtype} 
\usepackage{paracol}   
\usepackage{changepage} 
\usepackage{marvosym}
\usepackage{graphicx}
\usepackage{caption}
\usepackage[marginal]{footmisc}
\usepackage{float}
\usepackage{hyperref}
\hypersetup{
    colorlinks=true,
    linkcolor=blue,
    citecolor=green,
    urlcolor=red,
    pdftitle={Hyperlink Test}
}

\begin{document}
\begin{frontmatter}

\title{CKD-EHR:Clinical Knowledge Distillation for Electronic Health Records}
\author{Junke Wang\fnref{co-first}}
\ead{18236206585@163.com}
\author{Hongshun Ling\fnref{co-first}}
\ead{linghs0727@hotmail.com}
\author{Li Zhang\corref{cor}}
\ead{zhangli1996@163.com}
\author{Longqian Zhang}
\ead{13393769196@163.com}
\author{Fang Wang}
\ead{2022012@mail.scuec.edu.cn}
\author{Yuan Gao}
\ead{gaoyuan@scuec.edu.cn}
\author{Zhi Li}
\ead{lizhi@mail.scuec.edu.cn}

\cortext[cor]{Corresponding author}
\fntext[co-first]{Equal Contribution} 

\affiliation{organization={School of Medical Engineering,South-Central Minzu University},
            addressline={182 Minzu Avenue, Hongshan District}, 
            city={WuHan},
            postcode={430074}, 
            state={HuBei},
            country={China}}
        
\begin{abstract}
Electronic Health Records (EHR)-based disease prediction models have demonstrated significant clinical value in promoting precision medicine and enabling early intervention. However, existing large language models face two major challenges: insufficient representation of medical knowledge and low efficiency in clinical deployment. To address these challenges, this study proposes the CKD-EHR (Clinical Knowledge Distillation for EHR) framework, which achieves efficient and accurate disease risk prediction through knowledge distillation techniques. Specifically, the large language model Qwen2.5-7B is first fine-tuned on medical knowledge-enhanced data to serve as the teacher model. It then generates interpretable soft labels through a multi-granularity attention distillation mechanism. Finally, the distilled knowledge is transferred to a lightweight BERT student model. Experimental results show that on the MIMIC-III dataset, CKD-EHR significantly outperforms the baseline model: diagnostic accuracy is increased by 9\% (p<0.01), F1-score is improved by 27\%, and a 22.2 times inference speedup is achieved. This innovative solution not only greatly improves resource utilization efficiency but also significantly enhances the accuracy and timeliness of diagnosis, providing a practical technical approach for resource optimization in clinical settings. The code and data for this research are available at\url{https://github.com/209506702/CKD_EHR}.
\end{abstract}

\begin{highlights}
\item Proposes CKD-EHR, the first framework integrating data augmentation with knowledge distillation for EHR analysis.
\item Reduces computational costs via efficient teacher model fine-tuning while maintaining high accuracy.
\item Enhances medical prediction with a multi-label adaptive projection head mechanism for nuanced soft label learning.
\item Improves clinical decision support by dynamically constructing a medical knowledge base from real-world data.
\end{highlights}

\begin{keyword}
Large language models\sep disease prediction\sep model distillation\sep fine-tuning
\end{keyword}


\end{frontmatter}


\section{Introduction}
\label{sec1}

With the continuous progress of artificial intelligence technology, especially the development of Large Language Models (LLMs), their excellent data representation and understanding capabilities have significantly improved processing effectiveness and efficiency, offering new possibilities for automated decision support in the healthcare field\cite{xiao2024comprehensive}\cite{d2024large}. However, medicine heavily relies on specialized medical knowledge, and general-purpose large models often fall short in understanding and applicability when dealing with specialized medical information\cite{moor2023foundation}, making it difficult for them to conduct effective diagnosis or prediction. \par
To address this issue, researchers have introduced Retrieval-Augmented Generation (RAG) technology\cite{lewis2020retrieval}. By providing large language models with richer external knowledge, RAG effectively enhances model accuracy and reliability. For example, Kresevic et al.\cite{kresevic2024optimization} combined RAG with prompt engineering to increase model accuracy to 99 \%; Darren Edge et al. \cite{edge2024local}integrated the advantages of retrieval-augmented generation and graph-based methods to improve question-answering capabilities over private corpora. However, existing experiments mostly retrieve lengthy text information from the web, which is often filled with irrelevant content and dubious truthfulness\cite{yao2023react}, severely affecting model accuracy.\par
Moreover, due to their vast number of parameters and complex architectural design, large language models require substantial computational power for training and inference, and place extremely high demands on hardware facilities. When processing large-scale datasets, every step, from data preprocessing and cleaning to feature extraction, consumes a significant amount of time and computational resources\cite{samsi2023words}. Additionally, to achieve higher accuracy, these models typically require multiple rounds of iterative optimization\cite{rabgay2023multiple}, further extending the overall time cost. The inclusion of RAG exacerbates this situation. Therefore, how to reduce computational resource consumption and shorten computation time while ensuring model accuracy has become an important direction for current research.\par
In this paper, we propose a clinical knowledge distillation framework from Qwen2.5-7B to BERT. This technique integrates external clinical knowledge with Electronic Health Records (EHRs) to enhance the large model’s understanding of medical concepts. Specifically, we introduce an Efficacy-Aware Data Fusion (EADF) strategy that extracts and analyzes real-world medical knowledge from patient records—rather than retrieving generic text from the web—to ensure authenticity and relevance \cite{xu2024ram}.

We then apply Low-Rank Clinical Knowledge Distillation (LoRCKD), where the teacher model (Qwen2.5-7B) generates soft labels to guide the student model (BERT) via a LoRA-based fine-tuning process \cite{ko2024distillm}. While maintaining high prediction accuracy, this approach significantly reduces computational resource demands and speeds up inference.To enhance the effectiveness of knowledge transfer, we incorporate a Multi-Label Adaptive Projection Head (MLAPH) in the hidden layer of the teacher model to produce soft labels for multi-label disease prediction. These soft labels are richer than traditional hard labels, as they reflect the model’s full probability distribution over all potential outputs, helping to capture subtle uncertainties inherent in clinical decision-making \cite{ridnik2023ml}. During training, the student model learns not only from the teacher’s outputs but also directly from ground-truth EHR labels, ensuring both fidelity to the original data and alignment with distilled knowledge. This integrated approach enables the student model to inherit the performance advantages of the large model while maintaining efficiency and adaptability for downstream clinical tasks.\par
\noindent  Main contributions:\par
\begin{itemize}
  \item
  We propose a clinical knowledge-enhanced distillation framework (CKD-EHR) that, for the first time, deeply integrates data augmentation with knowledge distillation. By dynamically constructing a medical knowledge base from real-world clinical data and generating augmented samples through symptom-treatment association analysis, the framework significantly improves model accuracy and generalization in processing electronic health records (EHR), thereby enhancing the performance of clinical decision support systems.    
  \item
  We are the first to distill a large language model (LLM) into a BERT-based model and perform efficient fine-tuning of the teacher model. This approach maintains high prediction accuracy while substantially reducing computational resource requirements and inference latency.
  \item
  We introduce a multi-label adaptive projection head to generate soft labels, enabling the student model to better capture subtle distinctions in the medical domain and improve final prediction accuracy.

\end{itemize}\par
Through this method, we not only solve the problem of insufficient knowledge understanding of large models in the medical field but also significantly optimize the utilization efficiency of computational resources, providing new ideas for future medical AI applications.

\section{ Related work}
\label{sec2}
Large Language Models (LLMs) have demonstrated remarkable generalization capabilities in Natural Language Processing (NLP) tasks and have been widely applied in translation\cite{khanna2024artificial}, summarization\cite{van2024adapted}, sentiment analysis\cite{kornblith2025analyzing}, question-answering systems\cite{liu2023webglm}, and chatbots, achieving impressive results. However, when it comes to specialized knowledge domains, especially the healthcare industry, general LLMs often fall short. For instance, Zada T et al.\cite{zada2025medical} evaluated the effectiveness of LLMs in healthcare and found that the accuracy rate was only 31\%, indicating a high risk of incorrect medical advice that could mislead medical decision-making. Additionally, Shicheng Xu et al.\cite{xu2024unsupervised} pointed out that when dealing with long and complex text knowledge, LLMs require deeper domain-specific optimization of their architecture and training methods to better handle the complexities of specialized fields. This calls for higher resource consumption and longer time investment.

To address these challenges, researchers have proposed knowledge distillation methods to transfer knowledge from complex teacher models to lightweight student models, advancing technical innovations in knowledge acquisition mechanisms and distillation algorithm design to gradually expand their adaptability in vertical domains. For example, Xiaohan Xu et al.\cite{xu2024survey} explored methods such as knowledge generation\cite{jiang2025effective}, multimodal and structured knowledge distillation\cite{dong2025modality}, loss function optimization\cite{ko2024distillm}, cross-architecture and multi-teacher distillation\cite{wang2023mted}. Melanie Sclar's \cite{sclar2022referee}experiments demonstrated the feasibility of symbolic knowledge distillation in reference-free tasks, opening up new pathways in the field of knowledge distillation. Moreover, Sam Shleifer et al.\cite{shleifer2020pre} compared three methods for training teacher models—Shrink and Fine-Tune, Knowledge Distillation, and Pseudo-Labeling—providing valuable insights for selecting appropriate model compression strategies. However, although these methods have improved resource efficiency to some extent, they mostly use general models as teacher models, and if the teacher model performs poorly in specialized domains, this limitation will also be passed on to the student model.\par
To tackle the aforementioned issues, we propose the CKD-EHR method. This method enhances data by analyzing real data as external knowledge and, unlike general models, fine-tunes the model with high-quality data for supervision. Subsequently, we perform knowledge distillation to train a lightweight model. In this way, our lightweight model can learn the knowledge of the fine-tuned large model while significantly reducing resource consumption, thereby achieving more efficient and accurate performance in specialized domains.

\section{ Methodology}
\label{sec3}

\subsection{Overall Framework}
\label{subsec1}
\begin{figure}[H]
\centering
\includegraphics[width=1\textwidth]{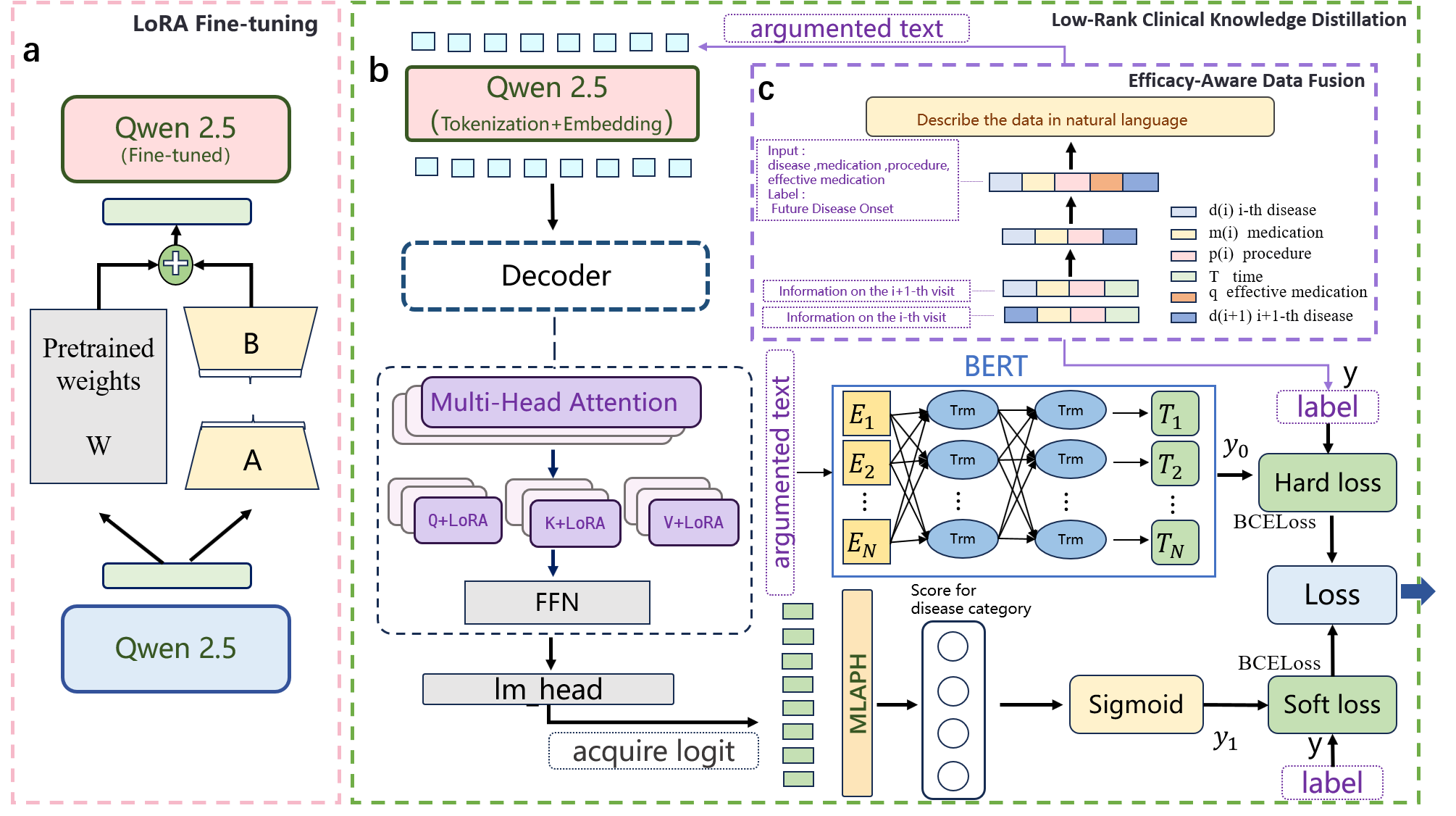}
\caption{The CKD-EHR framework includes:  (a) The basic principle of LoRA fine-tuning, (b)The low-rank clinical knowledge distillation  process, (c) The efficacy-aware data fusion process.
Here,  $y_{0}$ represents the prediction results of the student model,  $y_{1}$ represents the prediction results of the teacher model, and y is the true label of the data.}\label{fig1}
\end{figure}

We propose a new method called CKD-EHR for disease prediction tasks, which integrates three key components to enhance model performance. First, the Efficacy-Aware Data Fusion (EADF) module augments the training data by generating natural language descriptions based on clinical symptom-treatment associations from the MIMIC-III dataset, enriching the input context. Second, the Low-Rank Clinical Knowledge Distillation (LoRCKD) module leverages a medical-specialized large language model, Qwen2.5-7B-Instruct, as the teacher, which is fine-tuned using LoRA to adapt to the disease prediction task. The Multi-Label Adaptive Projection Head (MLAPH) is applied to the final hidden states of the teacher model to produce disease-specific probability scores. These predicted probabilities serve as soft labels. The student model learns by combining these soft labels with the true diagnosis results (hard labels). Together, these components improve the accuracy and robustness of disease prediction through data augmentation, knowledge distillation, and adaptive multi-label learning. The CKD-EHR framework is shown in Fig \ref{fig1}.

\subsection{Efficacy-Aware Data Fusion (EADF)}
\label{subsec2}

To enhance the performance of disease prediction models in real-world clinical scenarios, we propose an Efficacy-Aware Data Fusion (EADF) method. This approach focuses on constructing highly relevant and efficacy-informed data samples from high-quality medical records, providing causally-driven optimization for model training.

As illustrated in Fig \ref{fig1}(c), we first extract patient health records from the MIMIC-III database, selecting those diagnosed with any of 25 specific diseases. This forms the initial sample set $\{x_1, x_2, \dots, x_n\}$. Each data point $x$ represents core information from a single patient visit and is formally defined as:

\[
x \in [\text{ID}, d^{(i)}, m^{(j)}, p^{(k)}, T]
\]

where:
\begin{itemize}
    \item $\text{ID}$ denotes the patient identifier;
    \item $d^{(i)}$ represents the diagnosis at the $i$-th visit;
    \item $m^{(j)}$ refers to the $j$-th medication administered during the $i$-th visit;
    \item $p^{(k)}$ refers to the $k$-th procedure performed during the $i$-th visit;
    \item $T$ indicates the timestamp of the visit.
\end{itemize}

Based on the temporal order $T$, we pair each patient's $i$-th visit with their subsequent $(i+1)$-th visit to form a Visit Pair, thereby capturing disease progression and treatment response. Each pair is represented as:

\[
x \in [d^{(i)}, m^{(j)}, p^{(k)}, d^{(i+1)}]
\]

Here, $d^{(i+1)}$ reflects the disease status at the next visit, indicating potential treatment outcomes.

Subsequently, we apply statistical methods to analyze the changes in disease state across visit pairs, quantifying the efficacy of different combinations of medications and procedures. The outcome is a ranked list of effective treatments:

\[
Q \in \{q_1, q_2, \dots, q_m\}
\]

Each $q_l$ represents an efficacy ranking metric for a treatment option concerning a specific disease. In practice, we extract the top five treatments from $Q$ with the highest observed efficacy, forming a high-confidence set of causal associations between diseases and interventions.

The samples are then augmented with efficacy information as follows:

\[
x \in [d^{(i)}, m^{(j)}, p^{(k)}, q, d^{(i+1)}]
\]

Here, $q$ is selected from $Q$ and corresponds to the most statistically effective treatment related to the initial diagnosis $d^{(i)}$.

To enhance interpretability, we further convert each structured sample into a natural language description that includes the initial diagnosis at the first visit, the specific medications and procedures applied, and the treatment efficacy ranking based on real-world outcomes.

All generated descriptions undergo rigorous medical terminology validation and quality control to ensure accuracy, professionalism, and clarity. Meanwhile, the subsequent diagnosis $d^{(i+1)}$ is used as a supervision label to represent the actual clinical outcome, supporting training and evaluation of downstream disease prediction models (see Appendix Figure~\ref{fig5}).

Through this EADF process, we construct high-quality training data that integrates structural features, causal signals, and interpretable language, offering clinically meaningful supervision to learning algorithms.
\subsection{Low-Rank Clinical Knowledge Distillation (LoRCKD)}
\label{subsec3}
\textbf{Teacher Model Training and Optimization}:
We employ the generative medical large language model Qwen2.5-7B-Instruct as the teacher model. This model is specifically designed for the medical domain and contains rich medical knowledge and domain-specific language understanding capabilities. To further enhance its performance in the disease prediction task, we conduct supervised fine-tuning using the training data, as shown in Fig \ref{fig1}(a). We adopt the Low-Rank Adaptation (LoRA) technique during fine-tuning to efficiently adapt the model parameters.Specifically, in the self-attention module of each layer of the Transformer encoder, we freeze the original weight matrix W($W \in \mathbb{R}^{d \times k}$ ), and introduce two low-rank matrices $A \in \mathbb{R}^{r \times k}$ and  $B \in \mathbb{R}^{d \times r}$. Only these two low-rank matrices are trained, rather than directly adjusting the entire weight matrix. The adjusted weight matrices for the query, key, and value projections are defined as follows:
\begin{align}
W'_{Q} &= W_{Q} + B_{Q} A_{Q} \\
W'_{K} &= W_{K} + B_{K} A_{K} \\
W'_{V} &= W_{V} + B_{V} A_{V}
\end{align}\par
Here, W is the original weight matrix, which is frozen and not updated, while W' is the adjusted weight matrix. By doing so, the number of trainable parameters is significantly reduced to r×(d+k), effectively lowering the risk of overfitting and greatly enhancing the fine-tuning efficiency. This approach allows the model to retain the powerful feature extraction capabilities of the pre-trained model while precisely adapting it to the disease prediction task.\par
\textbf{Soft Label Generation and Knowledge Transfer}:
As shown in Fig \ref{fig1}(b), we input the augmented data into the teacher model for processing. The Tokenizer of the teacher model segments the data into tokens, transforming the raw text data into a form that machine learning models can understand. Subsequently, the Embedding layer maps these tokens into high-dimensional vectors (with a dimension of 2048), bringing words with similar meanings closer to each other in the vector space. The high-dimensional representations not only capture semantic relationships between tokens but also offer a rich expressiveness for modeling complex clinical semantics.\par
These embedded vectors are then processed through multiple Transformer encoder layers. The Transformer encoder refines the information of the input sequence by stacking multiple encoding layers. Each layer contains a Multi-head Self-Attention mechanism, which allows the model to simultaneously focus on information from different representation subspaces, thereby more effectively capturing long-range dependencies within the sequence. Subsequently, key layers of the teacher model are fine-tuned using LoRA, where low-rank matrix adjustments are applied within the self-attention modules to enhance the model's adaptability and efficiency.\par
Finally, a multi-label classification head is applied to project the model’s output onto the space of disease categories. For each data instance, the logits from the final layer are obtained through a forward pass of the teacher model. These logits, representing raw prediction scores for each label, are then converted into probabilities using the Sigmoid activation function.\par
The Sigmoid probability calculation formula is:
\begin{align}
p(y_i) = \frac{1}{1 + e^{-z_i}} 
\end{align}\par
The probability values after Sigmoid activation are the soft labels$y_{1}$. Soft labels not only reflect the teacher model's confidence in each category but also include estimates of uncertainty, providing richer information than hard labels. The resulting soft labels serve as learning targets for the student model, guiding it to mimic the teacher model's behavior and improve generalization, helping it better imitate the teacher model's behavior and thereby improve its performance.\par
To further optimize the training of the student model, we calculate the cross-entropy loss for both the soft labels and the hard labels. The student model receives the same input data as the teacher model and generates prediction results$y_{0}$. We compare$y_{0}$ and$y_{1}$with the actual diagnosis results y extracted directly from the real patient data to calculate the cross-entropy loss. Finally, we combine the soft label loss and the hard label loss to obtain the total loss, which is used to guide the further optimization and training of the student model.\par
Since this method deals with a multi-label binary classification task, the teacher model generates independent probabilities for each category as soft labels, while the hard labels use 0 and 1 encoding. Therefore, we choose to use the Binary Cross-Entropy With Logits Loss function (BCEWithLogitsLoss), which allows for separate loss calculation for each category and does not assume mutual exclusivity between categories, thus more accurately reflecting the gap between the model's predictions and the true labels. The calculation formula is as follows:
\begin{align}
l_n = - w_n \left[ t_n \log(\sigma(x_n)) + (1 - t_n) \log(1 - \sigma(x_n)) \right]
\end{align}\par

where $t_{n}$ is the target label, which can be 0 or 1, and  $\sigma(X_{n})$is the probability obtained through the sigmoid function. For hard labels, the loss for the positive class depends on $log \sigma(X_{n})$, and the loss for the negative class depends on  $log (1-\sigma(X_{n}))$ For soft labels, when $t_{n}$is a floating-point number between 0 and 1, the loss will be weighted according to the value of $ t_{n}$, reflecting the probability estimate given by the teacher model.\par
Finally, the overall training objective combines both hard and soft label supervision signals, formulated as:
\begin{align}
\text{Total-Loss} = \alpha L_{\text{hard}} + \beta L_{\text{soft}}
\end{align}\par
$\alpha $ and $\beta$ are weighting parameters ($\alpha+\beta=1$), used to adjust the impact of hard label loss and soft label loss on the total loss.

\subsection{Multi-Label Adaptive Projection Head (MLAPH)}
\label{subsec4}
After the final hidden layer of the teacher model, we introduce a linear classification head to better support downstream multi-label classification tasks. Traditional knowledge distillation methods typically rely on soft label distillation, where the student model mimics the teacher model’s probability distribution over the entire vocabulary. This output takes the shape:

\begin{equation}
\text{Logits} \in \mathbb{R}^{\text{[batch\_size, sequence\_length, vocab\_size]}}.
\end{equation}

However, this approach faces several significant limitations in practice. First, language models often have extremely large vocabularies—sometimes up to hundreds of thousands of tokens—which makes computing the KL divergence during distillation computationally expensive and inefficient. Second, the tokens in the vocabulary do not directly correspond to the semantic labels used in downstream tasks. This indirect mapping introduces semantic noise, which can severely impact the model’s ability to correctly identify disease labels, especially in medical scenarios. Third, the positions of the generated tokens are highly dynamic in generative models. The same label information might appear at different positions across different samples or inference passes, making it difficult for the student model to stably align with the teacher model's knowledge representations.

To address these issues, we propose the multi-label adaptive projection head (MLAPH). This module operates directly on the hidden states output by the teacher model, with shape:

\begin{equation}
\mathbf{H} \in \mathbb{R}^{\text{[batch\_size, sequence\_length, hidden\_size]}},
\end{equation}

and applies a linear projection to map the deep semantic features into a fixed-dimensional label space:

\begin{equation}
\hat{\mathbf{y}} = \text{Linear}(\text{Pooling}(\mathbf{H})) \in \mathbb{R}^{\text{[batch\_size, num\_labels]}}.
\end{equation}

This approach eliminates the need to reconstruct label distributions from vocab logits, thereby simplifying the knowledge transfer process. In multi-label medical prediction tasks, for example, the model can directly predict disease label probabilities from its semantic representations, without relying on the content or position of generated tokens.

This mechanism significantly improves computational efficiency during both training and inference by avoiding the overhead of high-dimensional vocabulary distributions. It also enhances task-level adaptability by directly aligning the model's semantic representations with the target label space. Because it relies on stable hidden representations rather than dynamic token outputs, MLAPH ensures more robust knowledge transfer. Furthermore, this structure offers strong flexibility and scalability, allowing it to naturally accommodate input sequences of varying lengths and different numbers of target labels.

Overall, MLAPH enables efficient alignment between the original language modeling objectives and the demands of multi-label classification, allowing generative large language models to perform more accurately, reliably, and efficiently in professional domains such as medical prediction.

\section{Experiments}
\label{sec4}
\subsection{Dataset Description}
\label{subsec1}
We selected the MIMIC-III (Medical Information Mart for Intensive Care III) dataset for our experiments.\cite{johnson2016mimic} The MIMIC-III dataset was constructed by the Laboratory for Computational Physiology at the Massachusetts Institute of Technology. It is a dataset composed of the health data information of over 40,000 patients. It includes demographic information, vital signs measurements taken at the bedside (approximately one data point per hour), laboratory test results, procedures, medications, nursing staff work records, imaging reports, and mortality information (both in-hospital and out-of-hospital). We used MIMIC-III to perform a phenotyping task \cite{harutyunyan2019multitask}, which is a combination of 25 separate binary classification tasks, predicting whether the patient will have the diseases shown in Fig \ref{app1} during their next visit based on their last visit information.

\subsection{Experimental Setup}
\label{subsec2}
We tested the CKD-EHR method for disease prediction based on patient information. The teacher model used was the Qwen2.5-7B-Instruct model with a learning rate set at 1.0e-5. The student model was the bert-base-uncased model, using the Adam optimizer with a learning rate of 2e-5, a batch size of 32, and 10 iterations.
\subsubsection{Baselines}
To verify the effectiveness of the CKD-EHR method, we compared the experimental results with the following baselines, all of which aim to improve EHR prediction performance through knowledge augmentation: Transformer \cite{li2020behrt}, GCT \cite{choi2020learning}, HYGT \cite{xu2023hypergraph}, MedRetriever \cite{ye2021medretriever}, SeqCare \cite{xu2023seqcare}, GraphCare \cite{jiang2023graphcare}, CORE \cite{van2021clinical}, BEEP \cite{naik2021literature}, and RAM-EHR \cite{xu2024ram}.We also employed ClinicalBERT\cite{unknown},BioBERT\cite{10.1093/bioinformatics/btz682} and Bio+ClinicalBERT\cite{Alsentzer2019PubliclyAC}, models specifically designed for processing clinical text.
\subsubsection{Evaluation Metrics}
To comprehensively evaluate the performance of the CKD-EHR method, we selected the following evaluation metrics:\par
\textbf{Accuracy}: Measures the proportion of correct predictions made by the model, which is the most intuitive performance metric.\par
\textbf{AUROC}: Used to assess the model's ability to distinguish between different categories in predictions.\par
\textbf{AUPR }: Particularly suitable for imbalanced datasets, it measures the model's precision performance at different recall rates.\par
\textbf{Macro-F1}: Calculates the average of the F1 scores for each category, without considering the number of samples between categories, which can more fairly evaluate the model's performance in multi-classification tasks.
\subsection{Experimental Results}
\label{subsec3}
\subsubsection{Main Inference Performance}

Table \ref{table1} shows the experimental results of CKD-EHR compared with the baseline methods. Since our method enhances the data using information associated with the dataset and incorporates soft labels generated by a fine-tuned large model, the student model achieves significant improvements in evaluation metrics compared with the baseline models. Specifically, there is an increase of approximately 10\% in accuracy, a 30\% improvement in F1 score, and enhancements of 4\% and 5\% in AUC and AUPR, respectively. These results demonstrate that augmenting the data with relevant external information and training the student model with soft labels generated by a fine-tuned large model can effectively boost model performance. Furthermore, CKD-EHR outperforms both RAG-based models and those specifically designed for processing clinical text.
\begin{table}[htbp]
\centering
{
\begin{tabular}{lcccc}
\toprule
\textbf{Model} & \textbf{ACC} & \textbf{F1} & \textbf{AUC} & \textbf{AUPR} \\
\midrule
Transformer\cite{li2020behrt} & 76.18 & 42.75 & 80.61 & 67.12 \\
GCT\cite{choi2020learning} & 77.20 & 37.57 & 78.62 & 64.87 \\
HyGT\cite{xu2023hypergraph} & 78.07 & 44.93 & 81.09 & 68.08 \\
MedRetriever\cite{ye2021medretriever} & 77.15 & 39.29 & 80.14 & 48.45 \\
SeqCare\cite{xu2023seqcare} & 77.44 & 39.16 & 80.98 & 68.53 \\
GrphCare\cite{jiang2023graphcare} & 80.11 & 44.33 & 82.26 & 71.19 \\
CORE\cite{van2021clinical} & 79.63 & 43.76 & 82.05 & 70.79 \\
BEEP\cite{naik2021literature} & 79.90 & 44.15 & 82.67 & 71.58 \\
RAM-EHR\cite{xu2024ram} & 85.54 & 53.01 & 87.14 & 71.76 \\
\midrule
BioBERT\cite{10.1093/bioinformatics/btz682} & 94.51 & 74.20 & 89.85 & 75.40 \\
ClinicalBERT\cite{unknown} & 94.62 & 76.78 & 90.19 & 75.84 \\
Bio+ClinicalBERT\cite{Alsentzer2019PubliclyAC} & 94.54 & 75.44 & 89.92 & 75.72 \\
CKD-EHR &\textbf{94.76} & \textbf{80.25} & \textbf{91.11} & \textbf{76.33}\\
\bottomrule
\end{tabular}
}
\caption{
Comparison of our method CKD-EHR with baseline models using the MIMIC-III dataset (\%). Higher values indicate better performance.}
\label{table1}
\end{table}

\subsubsection{Inference Performance}

In further performance evaluation, we conducted a detailed comparison of the efficiency of large language models (LLMs) and the CKD-EHR method during the data inference phase. To this end, we recorded the time consumption and memory usage of the two methods when processing a single piece of data, with the specific results shown in Table \ref{table2}. Comparative analysis revealed that the CKD-EHR method achieved a 22.2-fold improvement in time efficiency and a 34.96-fold optimization in memory usage compared to large language models. This significant performance enhancement fully demonstrates the CKD-EHR method's advantage in resource utilization efficiency, making it more feasible and scalable for practical applications.
\begin{table}[htbp]
\centering
\resizebox{\textwidth}{!}{ 
\begin{tabular}{lcc}
\toprule
\textbf{Model} & \textbf{Single-sample inference time (s)} & \textbf{GPU memory usage} \\
\midrule
Qwen2.5-7B & 0.1756 & 14.3GB \\
CKD-EHR & 0.0079 & 418.8MB \\
\bottomrule
\end{tabular}%
}
\caption{
For the performance evaluation of CKD-EHR, we calculated the memory usage of different models on a GPU and the inference time for a single sample.}\label{table2}
\end{table}
\subsection{Ablation Studies}
\label{subsec4}
\subsubsection{Comprehensive Ablation}
In Table \ref{table3}, we conducted a detailed comparative analysis of the impact of different model components. The results show that when predicting with a large language model (LLM) using an efficacy-aware augmented dataset, all performance metrics are improved compared to using the original dataset. This indicates that the augmented dataset can effectively enhance the model's predictive capabilities, further validating the rationality and effectiveness of the EADF strategy.\par
Moreover, the CKD-EHR method, after LoRCKD, demonstrates unique advantages in performance. Although its F1, AUC, and AUPR metrics are slightly lower than those of the large model, its accuracy (ACC) is very close to the large model, and in some cases, the ACC metric of the CKD-EHR method is even slightly higher than that of the large model. This result indicates that the proposed CKD-EHR method is very close to the large model in performance and even surpasses it in some key metrics.\par
This performance is of great significance for lightweight models. In practical applications, lightweight models often need to achieve efficient predictive performance under limited computational resources and storage conditions. The CKD-EHR method can maintain an accuracy rate close to or even better than that of the large model while significantly reducing resource consumption, showing great application potential and prospects in the field of lightweight models.
\begin{table}[htbp]
\centering
{ 
\begin{tabular}{lc|c|ccc}
\toprule
\textbf{EADF} & \textbf{LoRCKD} & \textbf{ACC} & \textbf{F1} & \textbf{AUC} & \textbf{AUPR} \\
\midrule
$\times$ & $\times$ & 94.58 & 81.61 & 90.62 & 75.76 \\
$\times$ & $\checkmark$ & 94.75 & 79.56 & 91.06 & 76.25 \\
$\checkmark$ & $\times$ & 94.75 & 82.94 & 91.39 & 76.51 \\
$\checkmark$ & $\checkmark$ & 94.76 & 80.25 & 91.11 & 76.33 \\
\bottomrule
\end{tabular}
}
\caption{
We conducted ablation studies on the key components of CKD-EHR, including efficacy-aware data fusion (EADF) and low-rank clinical knowledge distillation (LoRCKD).Here, $\times$ indicates that the component is not used, while $\checkmark$ indicates that the component is used. When all are $\times$ , it means we only use the Qwen model for disease prediction.}\label{table3}
\end{table}

\subsubsection{Different Probability Calculation Methods}

In addition, we also explored and compared other methods for obtaining soft label probabilities. In this study, we abandoned the traditional projection head mechanism and instead directly obtained soft label probabilities from the logits layer output. Specifically, we retrieved the IDs of all labels in the vocabulary and represented the final label probabilities based on the probabilities corresponding to these label IDs. Given the varying token lengths of different labels, we chose to average the probabilities of all relevant IDs as the final label probability. The relevant experimental results are shown in Table \ref{table4}.\par
However, during this process, we realized that the uncertainty in the position of label outputs might have some impact on the calculation of soft label probabilities. Therefore, we conducted further in-depth research. Based on the number of labels \(d+1(i)\) in the original data, we split the data into \(n\) parts, keeping the data input unchanged and only altering the final labels. Subsequently, we re-fine-tuned the teacher model so that it would only output the disease prediction with the highest probability. On this basis, we reacquired the soft label probabilities. Ultimately, the experimental results showed that the performance of the method of calculating soft labels after splitting the data was slightly different but similar to that when the data was not split (as shown in Table \ref{table4}). This is likely because the existing teacher model already has strong robustness and can to some extent overcome the challenges brought by the uncertainty in the position of label outputs.\par
In summary, although the improvement of the above methods is limited, this attempt provides a new idea and direction for the future calculation methods of soft labels, which is conducive to a deeper understanding of model behavior and how to better utilize soft label information, or to find more effective ways to enhance the quality of soft labels.
\begin{table}[htbp]
\centering
{ 
\begin{tabular}{lcccc}
\toprule
\textbf{Method} & \textbf{ACC} & \textbf{F1} & \textbf{AUC} & \textbf{AUPR} \\
\midrule
avg-prob & 94.76 & 80.01 & 91.06 & 76.29 \\
single-cls-prob & 94.75 & 79.96 & 90.99 & 76.24 \\
\bottomrule
\end{tabular}%
}
\caption{avg-prob: Calculate the average probability based on the ID of the label in the vocabulary.
single-cls-prob: Calculate probability values by splitting the data into multiple individual category labels.
}\label{table4}
\end{table}
\subsection{Experiments on Distillation $\alpha$ Parameter}
\label{subsec5}

\begin{figure}[H]
\centering
\includegraphics[width=1\textwidth]{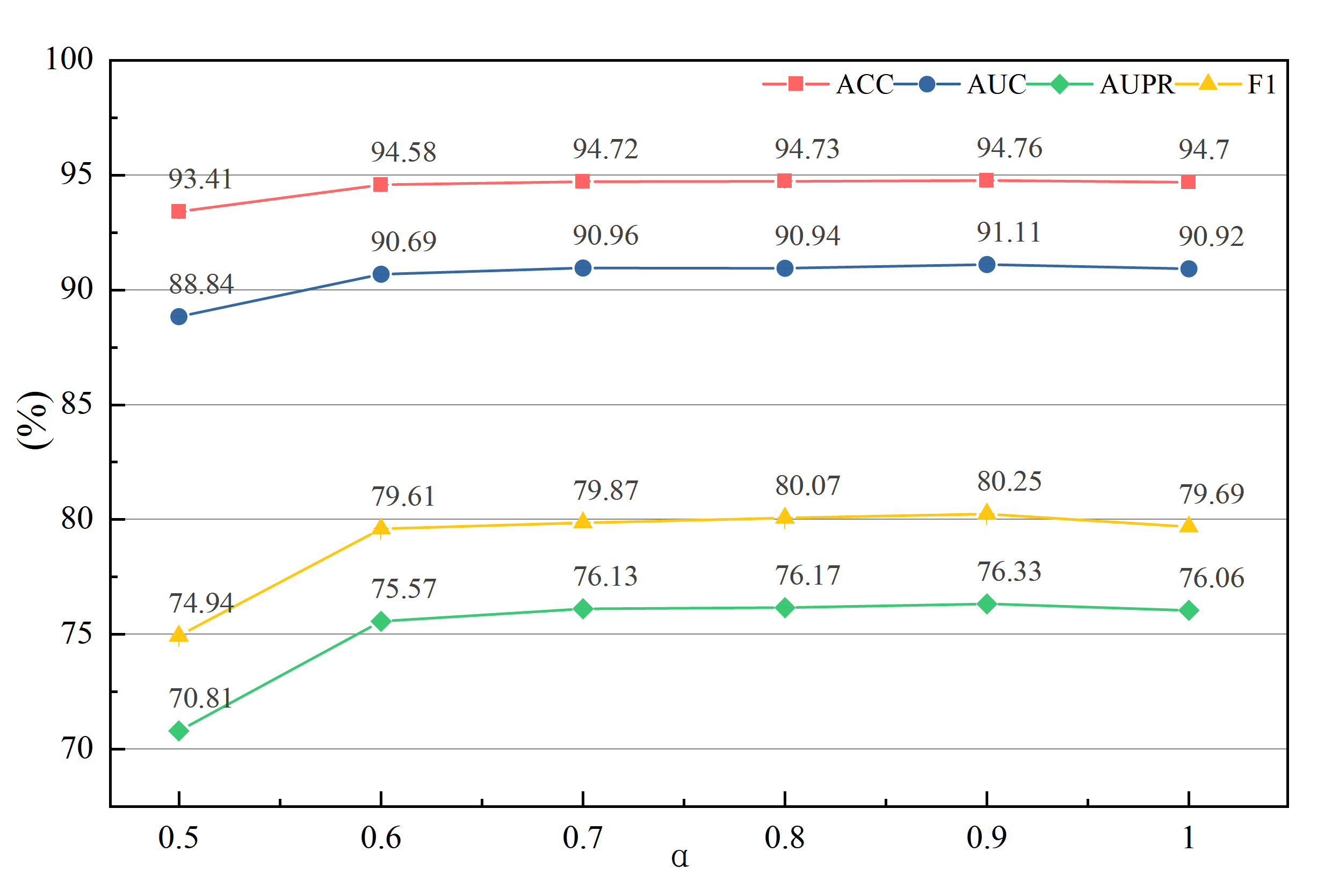}
\caption{Changes in indicators under different pre-soft target contribution rates $\alpha$
}\label{fig2}
\end{figure}

In the training of the student model, we assessed the impact of the parameter $\alpha $ on model performance. Below is the result Fig \ref{fig2} for different contribution rates of hard labels and soft labels ($\alpha $, $\beta$), where $\alpha $ is the contribution rate of hard labels and $\beta$ is the contribution rate of soft labels ($\alpha $+$\beta$=1). When $\alpha $=0.9, it means that 90\% of the label contribution for BERT training comes from hard labels and 10\% from soft labels. When $\alpha $=1, it indicates that the data labels are entirely from hard labels.

From the results in Fig \ref{fig2}, we can observe that when  $\alpha$ is set to 0.9, the CKD-EHR method achieves the best performance across all evaluation metrics, outperforming the scenario where only hard labels are used. This indicates that soft labels provide very useful information in the training of the student model, which helps to enhance the model's predictive performance. As  $\alpha$ decreases, the model's performance metrics gradually decline.

\subsection{Visualization Experiments}
\label{subsec6}

\begin{figure}[H]
\centering
\includegraphics[width=1\textwidth]{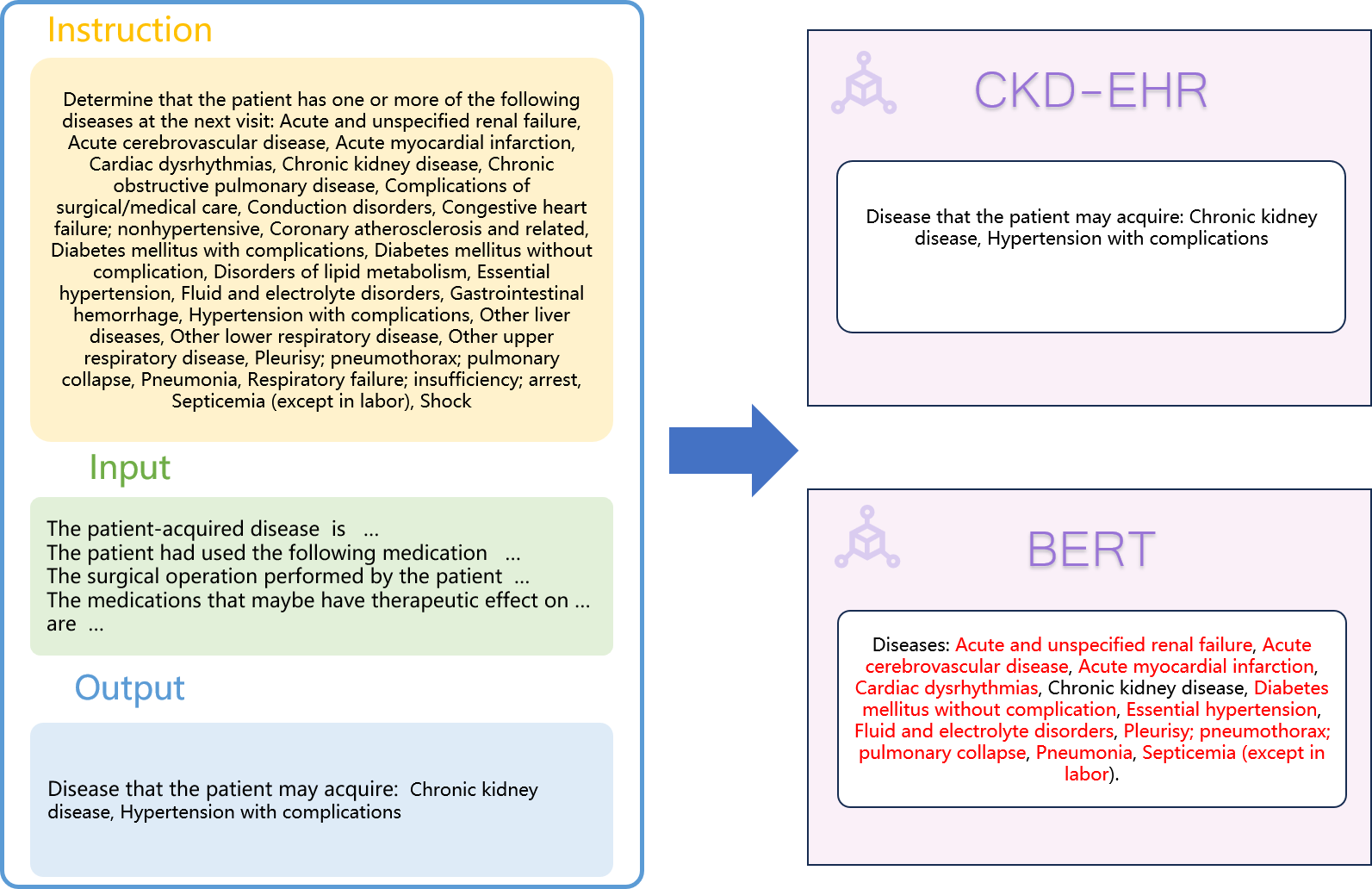}
\caption{Comparison of prediction results for the same data by CKD-EHR and BERT models. The left side shows the prompt, input, and output of the data, while the right side displays the prediction results of CKD-EHR and BERT, respectively. Black results indicate correct predictions, and red results indicate incorrect predictions.
}\label{fig3}
\end{figure}

We compared the disease prediction performance of different models on the same sample, with a fixed prediction threshold of 0.5 for all models.  Fig \ref{fig3} visually presents the prediction results of a representative case using both the BERT model and our proposed CKD-EHR method. As shown in the figure, the BERT model produces a large number of redundant predictions (highlighted in red), and it fails to predict some correct labels, demonstrating limitations in precision and recall. In contrast, the CKD-EHR method accurately identifies only the correct target diseases—Chronic kidney disease and Hypertension with complications—without introducing irrelevant labels. This indicates that CKD-EHR is more capable of capturing clinically relevant information while effectively reducing both false positives and false negatives.\par
\begin{figure}[H]
\centering
\includegraphics[width=1\textwidth]{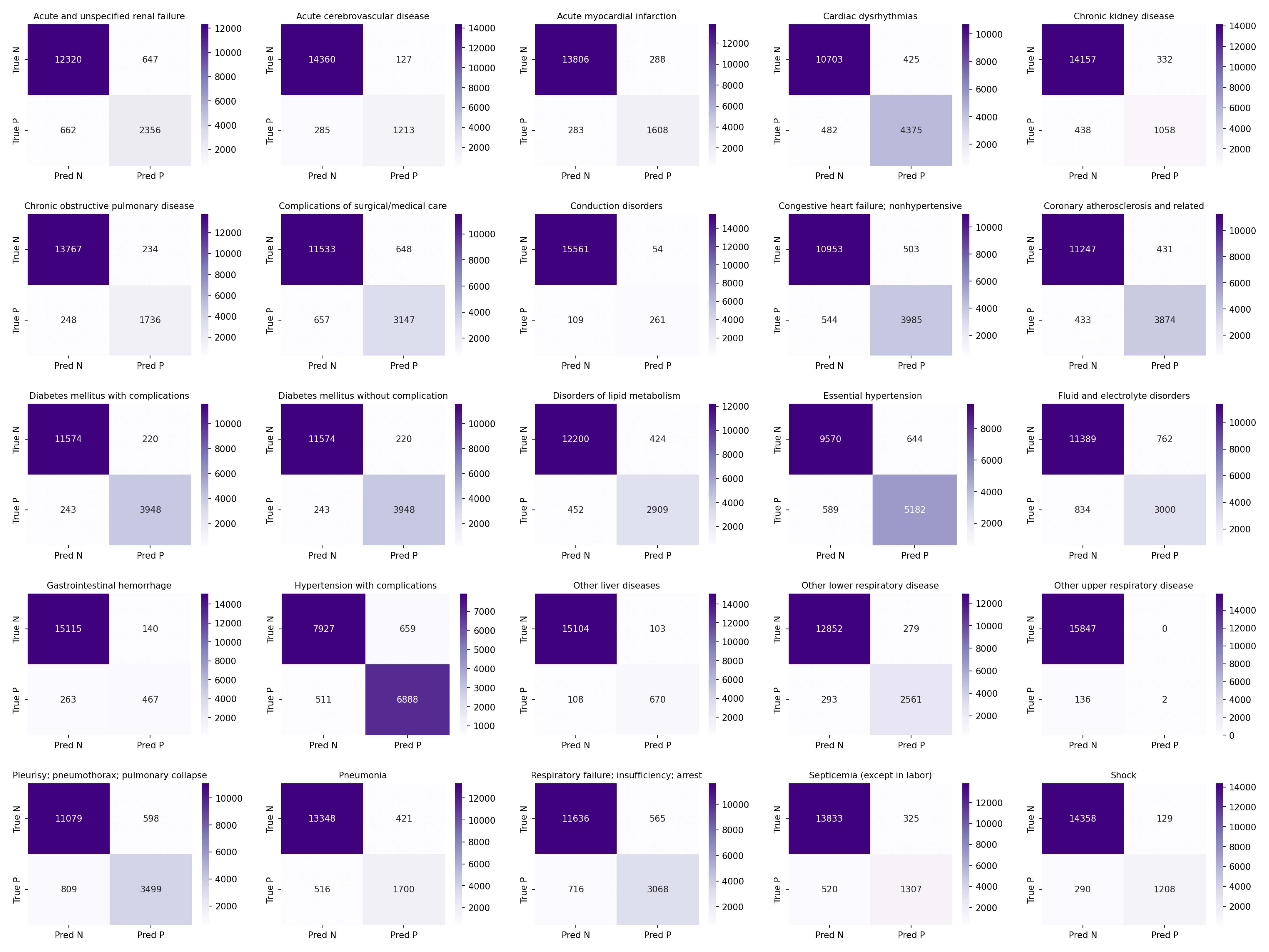}
\caption{Confusion experiments for different disease categories, with subplots showing TN, FP, FN, and TP for individual diseases, demonstrating the reliability of our model in the disease prediction task.
}\label{fig4}
\end{figure}

 Fig \ref{fig4} provides a comprehensive evaluation of the CKD-EHR model’s performance across 25 disease categories through a set of confusion matrices. Each subplot corresponds to one specific disease and shows the distribution of true positives (TP), false positives (FP), true negatives (TN), and false negatives (FN). The darker diagonal areas in most subplots reveal a high consistency between predicted and true labels, especially for high-incidence diseases like Essential hypertension and Cardiac dysrhythmias, confirming the model’s strong discriminative power.\par
However, for some low-support diseases, such as Other upper respiratory disease, there is a noticeable number of false negatives, likely due to data sparsity during training. Despite this, the model maintains high robustness even under multi-label and co-occurring disease conditions, demonstrating its potential as a reliable component in clinical decision support systems.

\section{ Conclusions}
\label{sec5}

In this paper, the CKD-EHR framework we proposed effectively addresses the dual challenges of insufficient knowledge representation and low resource efficiency of large language models in the medical field by deeply integrating clinical knowledge enhancement and knowledge distillation techniques. The CKD-EHR framework enhances data through efficacy-aware data fusion, dynamically constructing a medical knowledge base and improving model generalization via symptom-treatment correlation analysis. It also employs LoRA technology to efficiently fine-tune the large model, reducing parameter updates while retaining pre-trained knowledge. Moreover, the framework uses a Multi-Label Adaptive Projection Head (MLAPH) in conjunction with soft labels to convey the teacher model's uncertainty and to better capture inter-label dependencies in medical diagnosis to capture subtle differences in the medical domain, achieving a balance between performance and efficiency. Experiments show that on the MIMIC-III dataset, CKD-EHR increases diagnostic accuracy by 9\% and F1-score by 27\%, and optimizes memory usage by 34.96 times in resource-constrained scenarios, verifying its efficiency and reliability in such settings.\par

The results of the ablation experiments indicate that when certain components are removed, the performance of the CKD-EHR method drops slightly compared to the complete model, demonstrating the effectiveness and necessity of each part of the framework. In addition, the exploration of different probability calculation methods offers a new perspective for future improvements of soft labels.\par

Although CKD-EHR has shown advantages in a single dataset and a scenario with 25 specific diseases, its generalization ability still needs to be verified by extending it to multimodal medical data and a broader spectrum of diseases. Future research could explore cross-model distillation and dynamic weight adjustment strategies to enhance the framework's universality and robustness, providing stronger technical support for the clinical implementation of precision medicine and real-time decision support systems.
\section*{CRediT authorship contribution statement}
Junke Wang: Writing – original draft,Visualization, Validation,Data curation.
Hongshun Ling:Writing – review \& editing,Methodology,Supervision, Conceptualization.
Li Zhang: Writing – review \& editing,Supervision, Conceptualization,Project administration.
Longqian Zhang:Writing – review \& editing.
Fang Wang:Supervision, Project administration.
Yuan Gao:Supervision, Project administration.
Zhi Li:Supervision,Resources.

\section*{Declaration of competing interest}
The authors declare that they have no known competing financial interests or personal relationships that could have appeared to influence the work reported in this paper.
\section*{Acknowledgments}
This research is supported by Fundamental Research Funds for the Central Universities, South-Central Minzu University (No. CZQ24015), Key Research and Development Program of Hubei Province (No. 2022BAA037)
\section*{Data availability}

In this study, we used the MIMIC-III dataset from MIT’s Computational Physiology Laboratory, a widely recognized resource for healthcare data analysis. 
\appendix
\section{Phenotypes of 25 Diseases}
\label{app1}
In this study, we selected 25 representative diseases as prediction targets. By building accurate predictive models, our goal is to determine whether patients are likely to develop these diseases at their next clinical visit, thereby providing strong support for diagnosis and disease prevention.
\begin{table}[H]
\centering
\resizebox{\textwidth}{!}{ 
\begin{tabular}{lp{2cm}c}
\toprule
\textbf{Phenotype} & & \textbf{Type} \\
\midrule
Acute and unspecified renal failure & & acute \\
Acute cerebrovascular disease & & acute \\
Acute myocardial infarction & & acute \\
Cardiac dysrhythmias & & mixed \\
Chronic kidney disease & & chronic \\
Chronic obstructive pulmonary disease & & chronic \\
Complications of surgical/medical care & & acute \\
Conduction disorders & & mixed \\
Congestive heart failure; nonhypertensive & & mixed \\
Coronary atherosclerosis and related & & chronic \\
Diabetes mellitus with complications & & mixed \\
Diabetes mellitus without complication & & chronic \\
Disorders of lipid metabolism & & chronic \\
Essential hypertension & & chronic \\
Fluid and electrolyte disorders & & acute \\
Gastrointestinal hemorrhage & & acute \\
Hypertension with complications & & chronic \\
Other liver diseases & & mixed \\
Other lower respiratory disease & & acute \\
Other upper respiratory disease & & acute \\
Pleurisy; pneumothorax; pulmonary collapse & & acute \\
Pneumonia & & acute \\
Respiratory failure; insufficiency; arrest & & acute \\
Septicemia (except in labor) & & acute \\
Shock & & acute \\
\bottomrule
\end{tabular}
}
\caption{The 25 diseases predefined for the disease prediction task
}\label{table5}
\end{table}

\section{Prompt Design}
\label{app2}

To fully leverage the available data for effective prediction, we carefully designed a series of prompt templates to guide the model in understanding and generating more accurate predictions. The table below presents the data after natural language conversion and the corresponding prompt design. These carefully crafted prompts enable the model to analyze patient information from multiple perspectives, improving the accuracy and reliability of its predictions. In future research, we will continue to refine and optimize these prompts based on actual prediction outcomes to further enhance model performance.

\begin{figure}[H]
\centering
\includegraphics[width=1\textwidth]{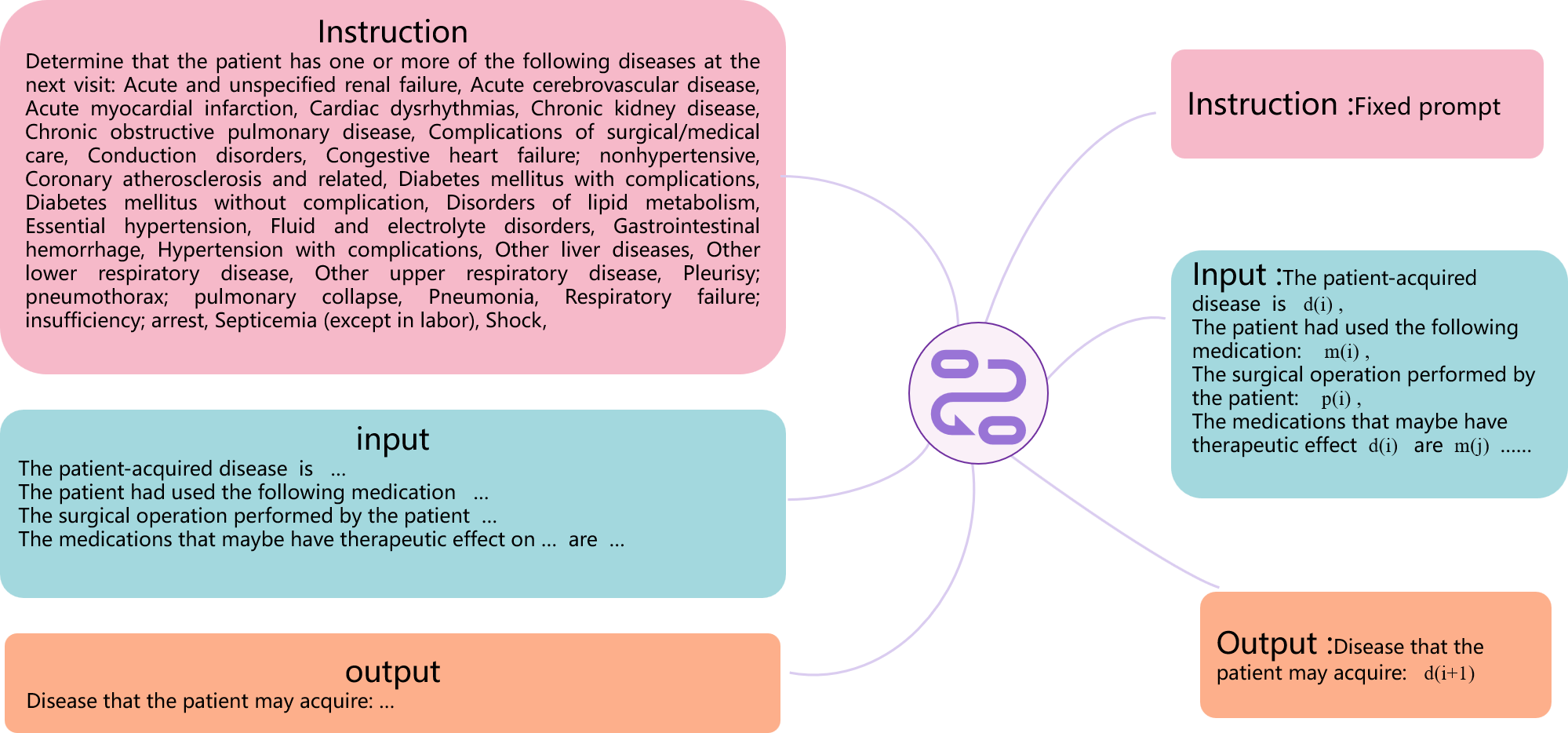}
\caption{Prompt design for the instruction, input, and output when feeding into the large model.
}\label{fig5}
\end{figure}

\bibliographystyle{elsarticle-num-names}
\bibliography{cas-refs}

\end{document}